\documentclass{article}
\usepackage{float}

 \usepackage[main, final]{neurips_2026}
\usepackage[utf8]{inputenc} % allow utf-8 input
\usepackage[T1]{fontenc}    % use 8-bit T1 fonts
\usepackage{hyperref}       % hyperlinks
\usepackage{url}            % simple URL typesetting
\usepackage{booktabs}       % professional-quality tables
\usepackage{amsfonts}       % blackboard math symbols
\usepackage{nicefrac}       % compact symbols for 1/2, etc.
\usepackage{microtype}      % microtypography
\usepackage{xcolor}         % colors

\usepackage{amsmath}
\usepackage{amssymb}
\usepackage{mathtools}
\usepackage{amsthm}

\usepackage[capitalize,noabbrev]{cleveref}
\usepackage{nicefrac}       
\usepackage{graphicx}
\usepackage{subcaption}
\usepackage{booktabs} 
\usepackage{xspace}

\theoremstyle{plain}

\theoremstyle{definition}

\theoremstyle{remark}

\title{A Geometric View of Counterfactual Behavior: Interaction of Boundary Proximity and Local Support}

\author{%
Ioanna Gemou \quad
Matteo Gamba \quad
Randall Balestriero \quad
Ritambhara Singh \\
Brown University \\
\texttt{\{ioanna\_gemou, matteo\_gamba, randall\_balestriero, ritambhara\}@brown.edu}
}

\begin{document}

\maketitle

\begin{abstract}
Counterfactual explanations seek small, semantically meaningful changes to an input 
that alter a model’s prediction, and are widely used to interpret and audit machine learning systems. 
% In modern vision, language, and multimodal models, inputs are mapped via pre-training to representations, where decision boundaries partition the embedding space.
In modern vision, language, and multimodal systems, pretrained encoders map inputs to representation spaces, and downstream classifier heads impose decision boundaries within those spaces.
As a result, the feasibility and distance of nearby counterfactuals depend on boundary placement relative to the data.
Yet models with similar predictive performance can differ substantially in whether such changes are achievable and how far representations must move.
This work examines this variation using a standardized local search probe across several pretrained encoders and linear classifier heads.
Results show that despite similar predictive performance, models differ substantially in their counterfactual behavior.
Under fixed representations, varying only the classifier head alters counterfactual outcomes while leaving predictive performance largely unchanged. 
This variation is explained by the interaction of decision-boundary proximity and local data support,
which jointly determine whether prediction changes are both feasible and lie in regions supported by the data, 
and can also improve counterfactual search within fixed models. 
Together, these findings identify counterfactual behavior as a distinct dimension beyond predictive performance 
and show that it can be altered without changing accuracy, 
with implications for model selection, robustness, and the reliability of counterfactual methods.\footnote{Code available at \url{https://github.com/igemou/counterfactual-geometry}.}
\end{abstract}

\section{Introduction}
Counterfactual explanations ask: \emph{can a model’s prediction be changed through a small, semantically meaningful change to the input?}~\cite{wachter2017counterfactual, Verma2024}
They provide an intuitive and actionable form of explanation, 
and help interpret and audit machine learning systems in domains such as healthcare and finance, 
where it is important to understand whether nearby alternatives could change a decision~\cite{Ustun_2019,Karimi21}.

Most prior work on counterfactual explanations has focused on tabular data and feature-space perturbations~\cite{wachter2017counterfactual, mothilal2020dice}. 
These methods typically formulate counterfactual generation as a local optimization problem:
given an input, they search for a small prediction-changing perturbation subject to task-specific feasibility constraints.
While such constraints are essential, they do not fully characterize counterfactual behavior:
a prediction-changing perturbation may not exist within the imposed locality budget, and boundary crossing may still lead to regions with weak data support.
Related observations in adversarial robustness show that predictions can often be flipped by small perturbations that do not correspond to realistic nearby alternatives~\cite{fawzi2017analysis, tanay2016boundarytiltingpersepectivephenomenon}. 
Thus, whether a nearby prediction change is feasible and meaningful depends not only on reaching a decision boundary,
but also on how that boundary is positioned relative to the data.

In modern machine learning systems, this issue becomes more pronounced because prediction often proceeds through learned representations:
a pretrained encoder maps inputs to an embedding space, and a task-specific classifier partitions that space into decision regions~\cite{devlin2019bertpretrainingdeepbidirectional, radford2021learningtransferablevisualmodels, Bengio}. 
Studying counterfactual behavior in this space is therefore natural:
the encoder determines the feature geometry of examples, while the classifier head determines how decision boundaries cut through that geometry.

At the same time, prediction changes in embedding space are not automatically meaningful alternatives in input space; they must remain close to regions supported by the data.
Since different models can organize data differently in representation space even when predictive performance is similar~\cite{damour2020underspecification, kornblith2019similarityneuralnetworkrepresentations}, this raises a central question:
\textit{when do learned representations admit locally supported prediction changes?}

\begin{figure}
    \centering
    \includegraphics[width=\linewidth]{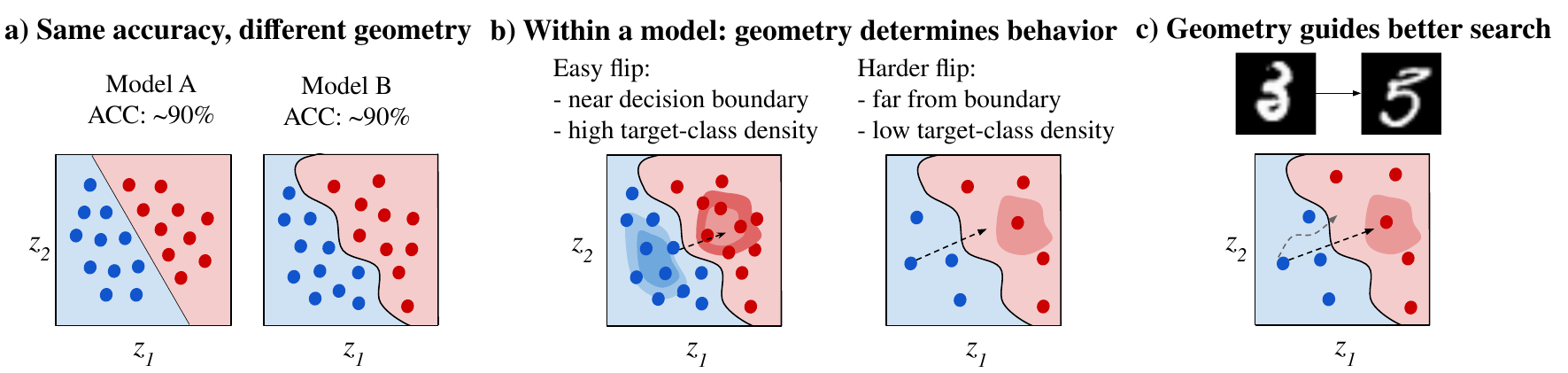}
    \caption{\textbf{Counterfactual behavior is shaped by boundary proximity and local data support.}
    Schematic in representation space. (a) Accuracy-matched models can differ in decision-boundary placement relative to the data. (b) Nearby prediction changes are easier for points closer to the boundary and to regions of higher target class support. (c) The same geometric signals can guide local search toward improved counterfactual outcomes under fixed constraints.}
\label{fig:overview}
\end{figure}

Motivated by these considerations, this work studies counterfactual behavior as a function of the interaction between local data support and classifier boundary placement.
\Cref{fig:overview}a shows that models with similar predictive performance can place decision boundaries differently relative to local data support.
To quantify these differences, in this work, a standardized local search probe evaluates models
under matched constraints and relates the behavior to two geometric quantities:
decision-boundary proximity (distance to the nearest boundary) and local data support (proximity to target class examples).
Across a comprehensive evaluation spanning several state-of-the-art vision, language, and multimodal models, this interaction consistently controls counterfactual outcomes.
Furthermore, under fixed representations, changing only the classifier head shifts the decision boundary and its relationship with local data support, altering whether prediction changes are feasible and supported without substantially changing predictive performance.
\Cref{fig:overview}b illustrates how these geometric quantities shape local behavior: 
points closer to the boundary and to regions of strong local data support are more amenable to prediction changes. 
Finally, \cref{fig:overview}c shows that the same geometric quantities can guide local search towards improved counterfactual outcomes.

\paragraph{Contributions.}
This work makes three contributions.
First, it introduces a geometric framework and standardized search probe for evaluating counterfactual behavior.
Second, it shows that under fixed embeddings, counterfactual behavior can vary substantially across classifier heads despite similar predictive performance, 
isolating the effect of decision-boundary placement.
Third, it shows that decision-boundary proximity alone is insufficient:
local data support and its interaction with boundary proximity predict whether nearby prediction changes are feasible and locally supported.
Together, these findings establish counterfactual behavior as a distinct axis beyond predictive performance and margin-based robustness alone, and show how it can inform model selection, robustness analysis, and the design of more reliable counterfactual explanation methods.

\section{Related Work}
This work is most closely related to four lines of research: 
representation geometry, decision-boundary analysis, counterfactual explanations and recourse, 
and model multiplicity. 
While these areas study related aspects of model behavior, the link between representation geometry and counterfactual behavior remains underexplored.

\paragraph{Geometry of learned representations.}
Learned representations often differ in how well classes separate and how data are organized in the embedding space.
Prior work has studied this relationship through the data-manifold framework, which characterizes separability using quantities such as manifold radius, intrinsic dimension, and inter-class correlations~\cite{Cohen2020, Chung_2018}. 
Subsequent work has studied the geometric structure in learned representations, including anisotropy, effective rank, and neural collapse~\cite{ethayarajh2019anisotropy,Royrank2007,papyan2020neural, kornblith2019similarityneuralnetworkrepresentations}. 
These approaches primarily focus on whether classes are separable or well-organized in representation space. 
In contrast, our work studies how representation geometry relates to whether nearby, locally supported prediction changes are reachable under fixed locality constraints.

\paragraph{Decision boundaries and adversarial robustness.}

Prior work has analyzed decision-boundary geometry through adversarial robustness, relating boundary crossing to margin, curvature, and local linearity~\cite{goodfellow2015explainingharnessingadversarialexamples, moosavidezfooli2016deepfoolsimpleaccuratemethod, fawzi2017analysis, tanay2016boundarytiltingpersepectivephenomenon, Elsayed2018largemargin,jiang2019predictinggeneralizationgapdeep}. 
These analyses focus on the minimal perturbation required to change a prediction, often independently of whether the perturbed point remains supported by the local data distribution.
Conversely, our work examines counterfactual changes that must both cross the decision boundary and remain in locally supported target regions, and therefore considers boundary proximity jointly with local data support.

\paragraph{Counterfactual explanations and recourse.}
Counterfactual explanation methods aim to construct minimal changes that alter model predictions~\cite{wachter2017counterfactual, poyiadzi2020face, mothilal2020dice}.
Latent-space and generative approaches improve plausibility by constraining perturbations to learned representations~\cite{poyiadzi2020face, Verma2024, Guidotti2022}, 
while algorithmic recourse studies whether actionable changes exist that can modify decisions~\cite{Ustun_2019, Karimi21, joshi2019realisticindividualrecourseactionable}.
Recent work also analyzes counterfactual explanations relating them to adversarial perturbations~\cite{pawelczyk2021exploringcounterfactualexplanationslens}.
This line of work focuses primarily on generating or analyzing counterfactuals for individual instances. Our work instead uses a standardized local search probe diagnostically to study counterfactual behavior as a property of models and representations.

\paragraph{Model multiplicity and underspecification.}
Models with similar predictive performance can exhibit materially different behaviors, including differences in explanations, robustness, and recourse~\cite{marx2020predictivemultiplicityclassification,fisher2019modelswrongusefullearning,damour2020underspecification}. 
Such variability can, in part, arise from differences in classifier boundary placement: retraining only the final layer can substantially alter behavior without modifying representations~\cite{kirichenko2023layerretrainingsufficientrobustness}. 
Our work complements this perspective by identifying the interaction of decision-boundary proximity and local data support as concrete geometric factors governing variation in counterfactual behavior.

\section{Geometric Framework for Counterfactuals}
\label{sec:framework}

Counterfactual explanations seek minimal, semantically meaningful changes that alter a model’s prediction.
Let $\mathcal X$ denote the input space, $\mathcal Z \subseteq \mathbb R^d$ the representation space, and $\mathcal Y$ the label space.
Given an input $x \in \mathcal X$, a pretrained encoder $e:\mathcal X \to \mathcal Z$ maps $x$ to a representation $z=e(x)$, and a classifier $f:\mathcal Z \to \mathcal Y$ predicts label $y=f(z)$.
Counterfactual generation is formulated in embedding space as

\begin{equation}
z'=\arg\min_{z' \in \mathcal Z} \|z'-z\|_2
\quad \text{s.t.} \quad
f(z') \neq f(z),
\end{equation}
following standard formulations~\cite{wachter2017counterfactual, poyiadzi2020face}. 
Under fixed optimization budgets and locality constraints, counterfactual success depends on whether a prediction-changing region can be reached within the local neighborhood of $z$ 
while remaining in regions supported by the data.
This perspective suggests two complementary geometric quantities: decision-boundary proximity and local target class support.

\paragraph{Decision-boundary proximity.}
The distance to the nearest decision boundary influences how easily a prediction can change. 
Points close to a competing class boundary should require smaller movement than points deep inside the current decision region.

Let $h(z)$ denote classifier logits, and let $y$ and $y'$ denote the top two predicted classes. Defining the logit gap
\begin{equation}
g(z)=h(z)_y-h(z)_{y'},
\end{equation}

a first-order approximation to the distance to the decision boundary is given by
\begin{equation}
d_{\mathrm{bd}}(z)\approx
\frac{|g(z)|}{|\nabla_z g(z)|_2+\varepsilon},
\label{eq:boundary_distance}
\end{equation}
following standard local linearization analyses~\cite{moosavidezfooli2016deepfoolsimpleaccuratemethod,fawzi2017analysis,tanay2016boundarytiltingpersepectivephenomenon,jiang2019predictinggeneralizationgapdeep,Elsayed2018largemargin}.
This provides a local proxy for the perturbation required to reach a decision boundary under a linear approximation of the classifier. 
Larger values indicate that the model requires greater local movement to achieve a prediction change.

\paragraph{Local target class support.}
Decision-boundary proximity alone does not determine counterfactual quality:
crossing a nearby boundary may lead to weak-support regions.
Motivated by manifold-aware counterfactuals~\cite{poyiadzi2020face} and algorithmic recourse~\cite{joshi2019realisticindividualrecourseactionable, Ustun_2019},
local target class support is measured using class-conditional neighborhoods in embedding space.
Let $\mathcal{R}_c$ denote a reference set of embeddings from class $c$, and let
\begin{equation}
r_k(z,c)=\operatorname{kNNdist}_k(z,\mathcal{R}_c)
\end{equation}
denote the Euclidean distance from $z$ to its $k$th nearest neighbor in $\mathcal{R}_c$.
In our counterfactual setting, support is evaluated with respect to the target class $y'$, and we write $r_k(z) := r_k(z,y')$ when the target class is clear from context.
Smaller values indicate stronger local data support, while larger values indicate that target-class points are farther away, reflecting lower local data support near $z$; more generally, larger nearest-neighbor distances correspond to sparser regions of the representation space~\cite{polianski2022}.
Here, the target-class score $r_k(z)$ provides an explicit, measurable notion of local data support that can be analyzed jointly with $d_{\mathrm{bd}}(z)$. 
Consequently, counterfactual behavior depends on the interaction between decision-boundary proximity and local data support.

\paragraph{Predictions for counterfactual behavior.}
Decision boundary proximity $d_{\mathrm{bd}}(z)$ and local target class support $r_k(z)$ jointly characterize counterfactual behavior across examples and across models with similar predictive performance.
Also, prediction changes should be easier when the decision boundary is close and local target class support is strong. 
This yields three geometric predictors: 
(a) counterfactual success decreases as $d_{\mathrm{bd}}(z)$ increases, 
(b) conditional on boundary proximity, counterfactual success further decreases as $r_k(z)$ increases, and 
(c) successful changes require larger movement and more optimization effort when either quantity is large. 
These predictors define testable relationships between embedding-space geometry and counterfactual outcomes.

Experiments test whether $d_{\mathrm{bd}}(z)$ and $r_k(z)$ jointly explain variation in counterfactual success, distance, and optimization effort across datasets, pretrained encoders, and classifier heads.
Unlike prior work that studies boundary geometry or local data support in isolation, 
this formulation makes their interaction explicit and tests its predictive power across models.

\section{Experimental Setup}\label{sec:setup}

Experiments compare several pretrained encoders using a standardized local search protocol, 
and vary classifier heads under fixed embeddings through repeated training 
with different initializations and regularization strengths.
The resulting behavior is then related to the two geometric quantities defined in \cref{sec:framework}.
Specifically, the geometric framework above motivates three hypotheses:

\begin{enumerate}
    \item \textbf{H1:} Models with similar predictive performance can nevertheless differ substantially in counterfactual success, perturbation distance, and optimization effort.
    
    \item \textbf{H2:} Under fixed embeddings, varying the classifier head can substantially change counterfactual behavior by shifting decision-boundary placement relative to the data, while leaving predictive performance largely unchanged.
    
    \item \textbf{H3:} Decision-boundary proximity and local data support jointly predict variation in counterfactual behavior.
\end{enumerate}

\paragraph{Datasets.}
Experiments span vision, medical imaging, text, and multimodal domains, covering synthetic structure, natural images, radiographs, language, and paired inputs to test whether the observed relationships hold across diverse representation settings.
\textit{Shapes}~\cite{rudman-etal-2025-forgotten} provides a controlled synthetic setting.  
\textit{MNIST}~\cite{lecun1998mnist} contains handwritten digit images across ten classes.  
\textit{Chest X-ray}~\cite{Kermany2018} consists of frontal radiographs labeled as \emph{pneumonia} or \emph{normal}.  
\textit{IMDb}~\cite{maasimdb} contains movie reviews labeled by sentiment.  
\textit{MM-IMDb}~\cite{aravelo2017} provides paired poster images and plot summaries; the original multi-label task is restricted here to binary classification for controlled comparison.

\paragraph{Representations and classifier heads.}
The evaluated encoders span supervised, self-supervised, and contrastive training paradigms. Vision encoders include ResNet-50~\cite{He_2016_CVPR}, ViT-B/16~\cite{dosovitskiy2021imageworth16x16words}, and DINOv2 ViT-B/14~\cite{oquab2024dinov2learningrobustvisual}. Language encoders include BERT~\cite{devlin2019bertpretrainingdeepbidirectional}, DistilBERT~\cite{sanh2020distilbertdistilledversionbert}, and RoBERTa~\cite{liu2019robertarobustlyoptimizedbert}. Multimodal encoders include CLIP~\cite{radford2021learningtransferablevisualmodels} and SigLIP2~\cite{tschannen2025siglip2multilingualvisionlanguage}.

All encoders are frozen, and embeddings are extracted from a predefined task representation layer:
global pooled features for vision models, pooled final hidden states for language models, and pretrained image–text embeddings for multimodal models. 
Multimodal representations differ from unimodal ones in that the embedding space is learned to align inputs across modalities rather than organize a single data distribution. 
Accordingly, proximity in this space does not correspond purely to class-conditional density, 
and local target class support should be interpreted as proximity to target class embeddings within this mixed, cross-modal structure.
A linear classifier head is trained on each representation, 
with validation-based model selection and held-out test evaluation. 
Unless otherwise stated, the primary probe is a regularized linear classifier.
% with linear SVMs as a maximum-margin baseline. 
This setup separates representation and boundary effects: encoders determine embedding geometry, while classifier heads determine decision boundaries. 
Variation across encoders reflects representation differences, whereas variation across heads reflects boundary differences on shared embeddings.

\paragraph{Geometry estimation.}
Geometric quantities are computed in representation space as described in \cref{sec:framework}. Decision-boundary proximity ($d_{\mathrm{bd}}(z)$) is estimated using a first-order linearization of the logit gap (eq. \ref{eq:boundary_distance}). Local data support ($r_k(z)$) is approximated by the distance to the $k$th nearest neighbor of the target class in representation space, with $k$ fixed across experiments ($k=20$ unless otherwise stated, see Appendix~\ref{app:density} for sensitivity analysis). Throughout, the target class is defined as the highest-scoring alternative class under the classifier.

\paragraph{Counterfactual evaluation protocol.}

All main experiments use a standardized local search probe in representation space.
For an embedding $z_0$ with predicted class $y$, the target class $y'$ is chosen as the highest-scoring alternative class.

Projected gradient updates are performed as
\begin{equation}
z_{t+1}
=
\Pi_{\|z-z_0\|_2\le\tau}
\left(
z_t-\eta \nabla_{z_t}\ell(z_t)
\right),
\label{eq:pgd_update}
\end{equation}
where $\ell$ is a target class objective, $\eta$ is the step size, and $\tau$ defines the trust region.

Optimization stops when the predicted label changes or a fixed iteration budget is reached. This procedure is intended as a standardized probe of whether a decision change is reached within a fixed local region and search budget, rather than a globally optimal counterfactual solver.

All optimization parameters ($\tau$, $\eta$, and budget) are fixed within each dataset and shared across models. 

\paragraph{Evaluation metrics.}
Let $z_i$ denote the original embedding for the $i$th example, and let $z_i'$ denote the final point reached by the projected-gradient local search procedure in Eq.~\eqref{eq:pgd_update}.

\textbf{Counterfactual success (CF-Suc)} is the fraction of examples for which the predicted label changes, where $N$ denotes the number of evaluated examples:
\begin{equation}
\mathrm{CF\text{-}Suc}
=
\frac{1}{N}
\sum_{i=1}^{N}
\mathbf{1}\{f(z_i') \neq f(z_i)\}.
\label{eq:cf_suc}
\end{equation}

\textbf{Counterfactual distance (CF-Dist)} is the average perturbation magnitude:
\begin{equation}
\mathrm{CF\text{-}Dist}
=
\frac{1}{N}
\sum_{i=1}^{N}
\|z_i' - z_i\|_2.
\label{eq:cf_dist}
\end{equation}

\textbf{Optimization effort (OptEff)} is the average number of iterations required to reach a label change, or $T$ if unsuccessful:
\begin{equation}
\mathrm{OptEff}(z_i)
=
\begin{cases}
t_i^* & \text{if } f(z_i^{(t_i^*)}) \neq f(z_i), \\
T & \text{otherwise},
\end{cases}
\label{eq:opteff_point}
\end{equation}
\begin{equation}
\mathrm{OptEff}
=
\frac{1}{N}
\sum_{i=1}^{N}
\mathrm{OptEff}(z_i).
\label{eq:opteff}
\end{equation}

\section{Results}

This section evaluates three hypotheses: 
(H1) accuracy-matched models differ in counterfactual behavior, 
(H2) behavior can be altered under fixed embeddings by varying the classifier head, and 
(H3) decision-boundary proximity and local data support explain this variation. 
Across domains, the results support all three, and an additional case study illustrates how these metrics can guide counterfactual search.

\paragraph{Models matched on predictive metrics can differ substantially in counterfactual behavior (H1).}

\Cref{tab:main_gaps} compares pairs of models with the \textit{smallest accuracy differences} within each dataset. 
Even when accuracy gaps are minimal at the reported precision, counterfactual behavior can differ sharply.

\begin{table}[ht]
\centering
\small
\caption{\textbf{Closest-accuracy encoder pairs within each dataset.} Despite small differences in predictive performance, paired models often exhibit large differences in counterfactual success (CF-Suc), counterfactual distance (CF-Dist), and optimization effort (OptEff). Reported values show mean differences with standard deviation across five seeds.}
\begin{tabular}{l l l r r r r}
\toprule
Dataset & Model A & Model B & ACC Gap & CF-Suc Gap & CF-Dist Gap & OptEff Gap \\
\midrule
Shapes    
& ResNet50 & ViT     
& $0.01 \pm 0.007$ 
& $0.53 \pm 0.009$ 
& $0.38 \pm 0.007$ 
& $56.55 \pm 1.720$ \\

IMDB      
& BERT & RoBERTa 
& $0.00 \pm 0.003$ 
& $0.19 \pm 0.009$ 
& $0.49 \pm 0.007$ 
& $43.75 \pm 1.409$ \\

MNIST     
& ResNet50 & ViT     
& $0.01 \pm 0.001$ 
& $0.65 \pm 0.002$ 
& $0.23 \pm 0.001$ 
& $46.72 \pm 0.054$ \\

ChestXray 
& ResNet50 & ViT     
& $0.01 \pm 0.029$ 
& $0.27 \pm 0.044$ 
& $0.16 \pm 0.035$ 
& $29.39 \pm 3.096$ \\

MM-IMDb   
& CLIP     & SigLIP2 
& $0.02 \pm 0.002$
& $0.15 \pm 0.004$ 
& $0.11 \pm 0.003$ 
& $6.34 \pm 1.047$  \\
\bottomrule
\end{tabular}
\label{tab:main_gaps}
\end{table}

On MNIST, for example, two models with an accuracy gap of 0.01 exhibit a 65-percentage-point gap in prediction flip rate (CF-Suc). 
The same pairs also differ in CF-Dist and OptEff,
indicating that similarly accurate models may require very different perturbation magnitudes and optimization effort to reach a counterfactual change. 
Similar discrepancies appear across vision, language, and multimodal settings, though the multimodal case shows smaller gaps overall. 
This likely reflects its representation geometry, 
which is shaped by cross-modal alignment and changes how local neighborhoods and target class support are structured.
This phenomenon is not limited to accuracy: repeating the analysis using validation cross-entropy loss yields the same conclusion, 
with small predictive-loss gaps still coinciding with large counterfactual gaps (Appendix~\ref{app:loss}). 

Together, these findings indicate that counterfactual behavior is not fully characterized by predictive accuracy or loss. 
They further suggest that it reflects how representations organize class structure through the relative placement of decision boundaries and locally supported target regions, providing a complementary diagnostic beyond standard metrics and motivating geometric characterizations for robustness and recourse.

\paragraph{Effective counterfactuals require local target class support beyond boundary crossing.}
Crossing a decision boundary does not guarantee reaching a locally supported target region. This analysis shows a clear separation between prediction changes that reach supported regions and those that do not, consistent with local target class support in~\cref{sec:framework}.
\begin{figure}[h]
    \centering
    \includegraphics[width=0.8\linewidth]{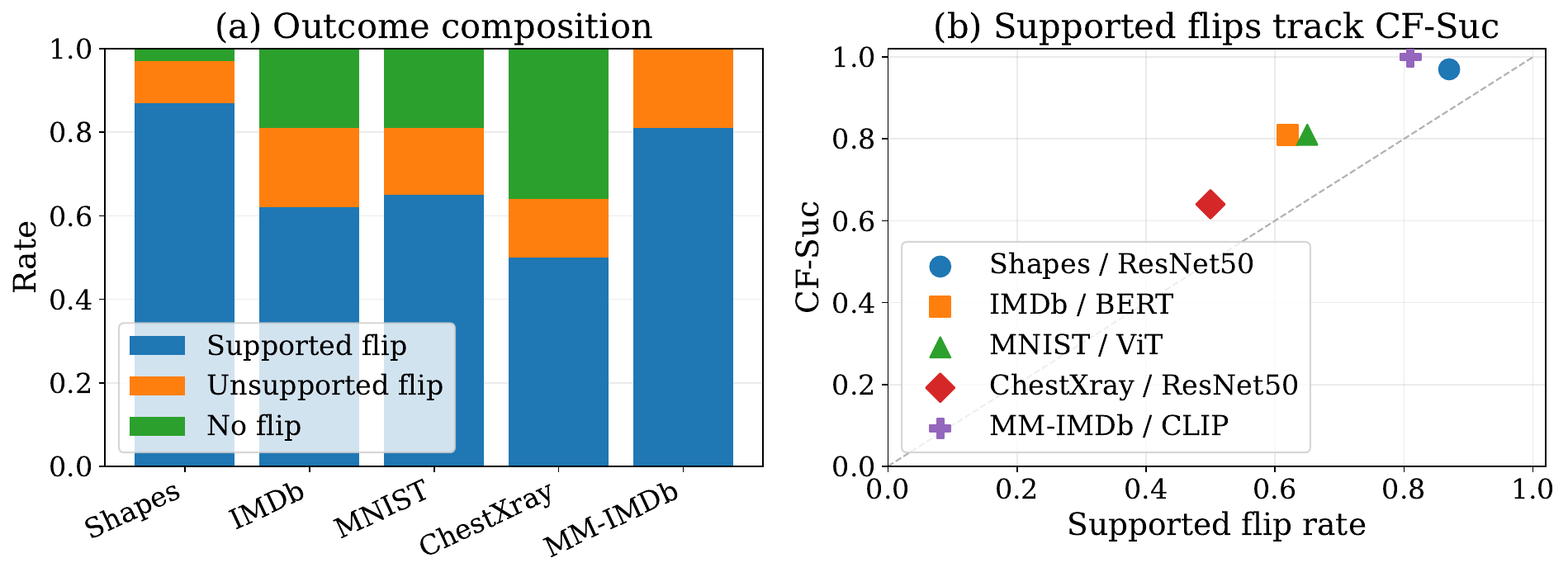}
    \caption{\textbf{Not all prediction changes are supported counterfactuals.}
    (a) Outcomes are partitioned into supported flips, unsupported flips, and no-flip cases. Supported flips reach target regions with sufficient local data support, while unsupported flips do not. (b) Supported-flip rate closely tracks CF-Suc, indicating that effective counterfactual behavior depends on reaching supported regions rather than merely crossing a decision boundary.}
    \label{fig:supported_flips}
\end{figure}
To examine this, search outcomes are partitioned into three categories: \emph{supported flips}: prediction changes that reach supported target regions, \emph{unsupported flips}: prediction changes that terminate in weak-support regions, and \emph{no-flip} cases: no prediction change within the budget. 
For counterfactual explanations, the most desirable outcomes are \emph{supported flips}.

\Cref{fig:supported_flips} summarizes these outcomes across representative models. 
Models differ not only in overall counterfactual success, but also in the fraction of successful changes that are locally supported. 
Across datasets, supported-flip rate closely tracks CF-Suc, indicating that effective counterfactual behavior depends on reaching supported target regions rather than merely crossing a decision boundary.
ChestXray exhibits a higher fraction of no-flip outcomes and lower supported-flip rates,
indicating that locally supported target regions are harder to reach. 
This likely reflects the structure of radiographic data, 
where class differences are subtle and representations are less cleanly separated, 
leading to weaker alignment between nearest-neighbor neighborhoods and target class regions.

Overall, these results show that boundary crossing is insufficient: effective counterfactual behavior requires reaching supported regions, reflecting the joint role of boundary proximity and local data support.

\paragraph{Counterfactual behavior can be altered through classifier boundary geometry (H2).}
The previous results show that counterfactual behavior depends on boundary crossing and local support. 
This raises the question of whether these relationships are purely descriptive, or whether counterfactual behavior can be controlled by modifying the classifier boundary while keeping the representation fixed.

\begin{figure*}[h]
    \centering
    \includegraphics[width=\textwidth]{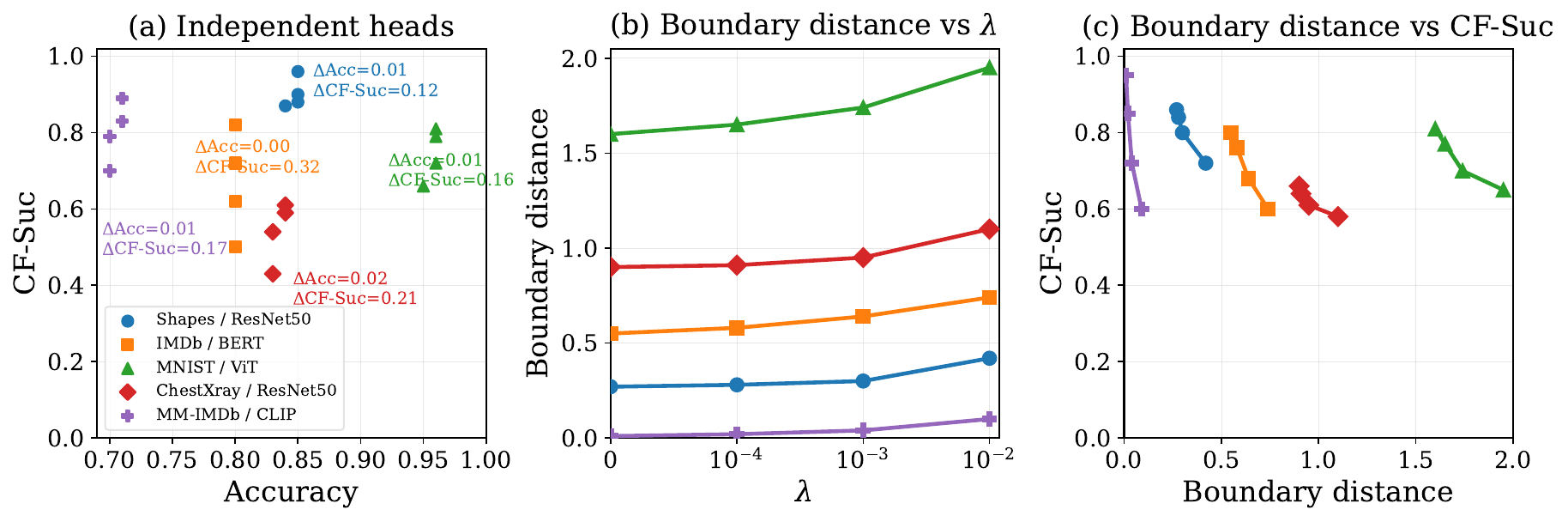}
    \caption{\textbf{Counterfactual behavior can be systematically altered through classifier boundary geometry.}
    A pretrained encoder remains fixed while the linear classifier head varies. (a) Independently trained heads on the same embeddings differ substantially in CF-Suc despite similar accuracy. (b) Sweeping regularization strength $\lambda$ systematically shifts mean boundary distance. (c) Larger boundary distance is consistently associated with lower CF-Suc across datasets. Counterfactual behavior therefore depends on classifier boundary placement as well as the representation.}
    \label{fig:fixed_repr_boundary}
\end{figure*}

To test this hypothesis, 
the setup freezes a pretrained encoder to produce a single set of embeddings
and trains multiple linear classifier heads on these embeddings using identical training protocols. 
Across runs, the representation, data, loss, optimizer, and hyperparameters remain fixed, 
while only initialization varies, affecting the learned decision boundary.
\Cref{fig:fixed_repr_boundary}a shows substantial variation in CF-Suc across independently trained heads despite relatively stable predictive accuracy.
The much larger vertical spread in CF-Suc than horizontal spread in accuracy indicates that classifier boundary placement
can materially affect counterfactual behavior  (\cref{tab:h2_appendix_ranges} quantifies this variation).

To assess whether this variation is systematic, 
we sweep probe weight decay while keeping the representation unchanged. 
Increasing regularization shifts mean boundary distance across datasets (\Cref{fig:fixed_repr_boundary}b). \Cref{fig:fixed_repr_boundary}c shows a clear monotone relationship: larger boundary distance corresponds to lower counterfactual success, 
while predictive accuracy remains nearly unchanged.

We also compare learned probes to linear SVMs trained on the same embeddings.
Because linear SVMs recover maximum-margin separators, 
they provide a simple reference for classifier boundary placement. 
Across datasets, probes that deviate more from the SVM boundary tend to exhibit lower CF-Suc and larger CF-Dist, 
further supporting the role of classifier geometry (Appendix~\ref{app:svm}).

Together, these results show that counterfactual behavior can be altered through classifier-induced boundary geometry.
This sensitivity reveals a brittleness of counterfactual explanations as model-auditing tools:
under fixed representations and similar predictive performance, retraining or regularizing only the classifier head can substantially change which prediction changes appear feasible and supported.
Thus, counterfactual behavior is not invariant across functionally similar models, and it depends critically on how the final classifier places boundaries relative to local data support.

\paragraph{Boundary proximity and local support jointly predict counterfactual behavior (H3).}
The intervention results above isolate the effect of classifier boundary placement. 
This motivates testing whether decision-boundary proximity alone explains variation,
or whether incorporating local target-class support provides additional predictive power.
\begin{figure}[h]
    \centering
    \small
        \includegraphics[width=0.9\linewidth]{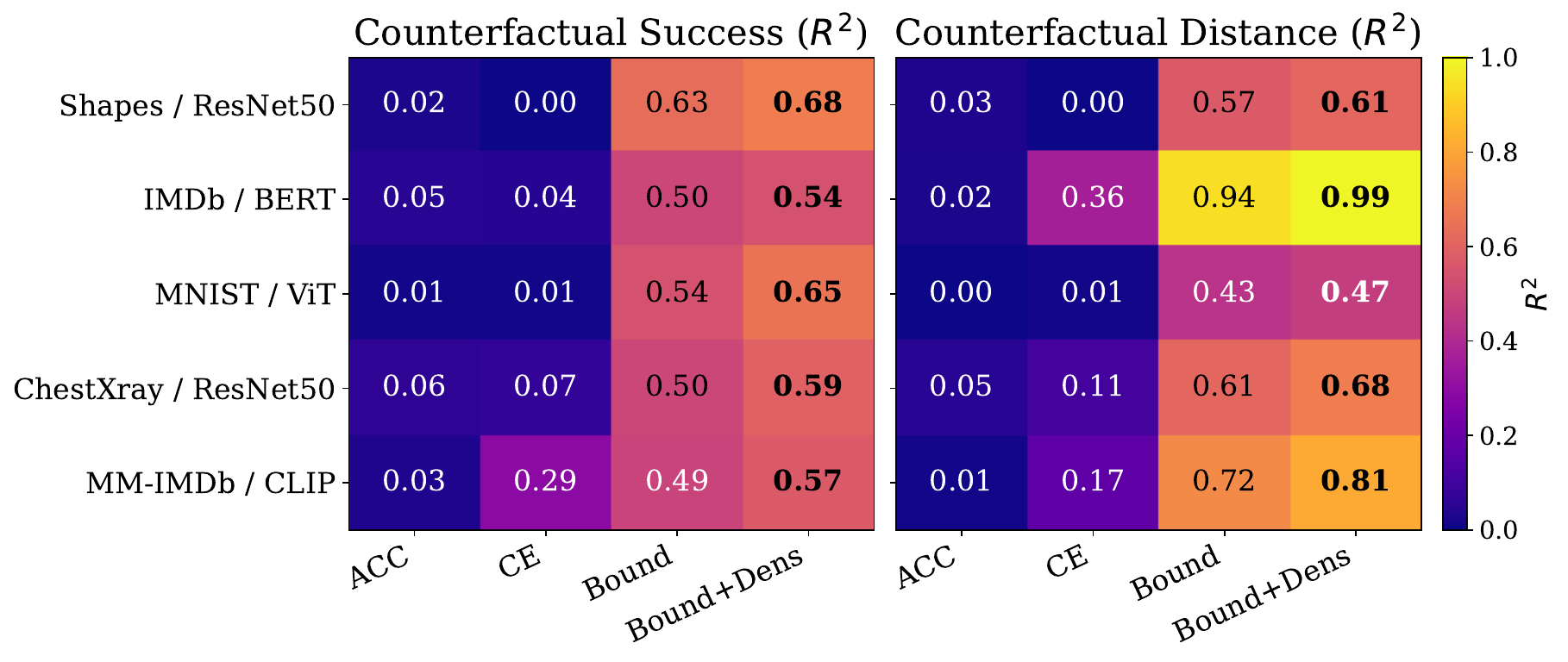}
    \caption{\textbf{Geometry predicts counterfactual behavior.}
    Held-out $R^2$ values for models predicting counterfactual success (left) and counterfactual distance (right). Standard predictive metrics (accuracy and cross-entropy loss) explain limited variance, whereas decision-boundary proximity is substantially more informative. Adding local data support yields further gains across datasets.}    \label{fig:regressions}
\end{figure}

To test this, predictive models are fit post hoc using decision-boundary proximity $d_{\mathrm{bd}}(z_i)$ and local data support $r_k(z_i)$, with evaluation on held-out examples.
\Cref{fig:regressions} summarizes predictive performance across datasets.
Across domains, decision-boundary proximity is consistently informative, 
and incorporating local data support yields further gains. 
Standard predictive metrics, including accuracy and cross-entropy loss, 
explain substantially less variance. 
These results are robust to the density estimator:
varying neighborhood size ($k \in \{5,10,20,50\}$) yields consistent trends 
and similar model rankings (Appendix~\ref{app:density}).

Overall, these findings show that counterfactual behavior is best explained by combining boundary proximity with local support, motivating geometry-aware analysis and counterfactual search.

\paragraph{MNIST case study: geometry priors can improve counterfactual search within the same model.}
\label{par:priors}

This case study shows how the geometric quantities above can guide counterfactual search. 
Unlike the main experiments, it augments the objective with geometry-aware priors that encourage prediction changes while remaining close to the original representation and moving toward locally supported target regions.
Let $y'$ denote the runner-up class and $\mathcal{R}_{y'}$ the corresponding set of reference embeddings. 
The optimization objective is
\begin{equation}
\ell(z_t)=
\ell_{\mathrm{clf}}(z_t)
+\lambda_{\mathrm{shift}}\|z_t-z\|_2
+\lambda_{\mathrm{knn}}\ell_{\mathrm{knn}}(z_t),
\end{equation}
where $\ell_{\mathrm{clf}}$ promotes a prediction change, $\|z_t-z\|_2$ penalizes large deviations, and $\lambda_{\mathrm{shift}}, \lambda_{\mathrm{knn}} > 0$ control their strength. 
The support term $\ell_{\mathrm{knn}}(z_t)$ penalizes distance to nearby target-class embeddings:
\begin{equation}
\ell_{\mathrm{knn}}(z_t)=
\frac{1}{k}\sum_{u\in \mathrm{kNN}_k(z_t;\mathcal{R}_{y'})}\|z_t-u\|_2^2.
\end{equation}
Step size $\eta$ and trust-region radius $\tau$ are scaled by dataset-level geometry statistics, with $\eta \propto s$ and $\tau \propto b$, where $s=\operatorname{median}_{x\in\mathcal V} r_k(e(x))$ is the median target-support radius and $b=\operatorname{median}_{x\in\mathcal V} d_{\mathrm{bd}}(e(x))$ is the median boundary proximity. The optimization budget and locality constraints are unchanged.
\begin{figure}[t]
    \centering
    \includegraphics[width=0.85\linewidth]{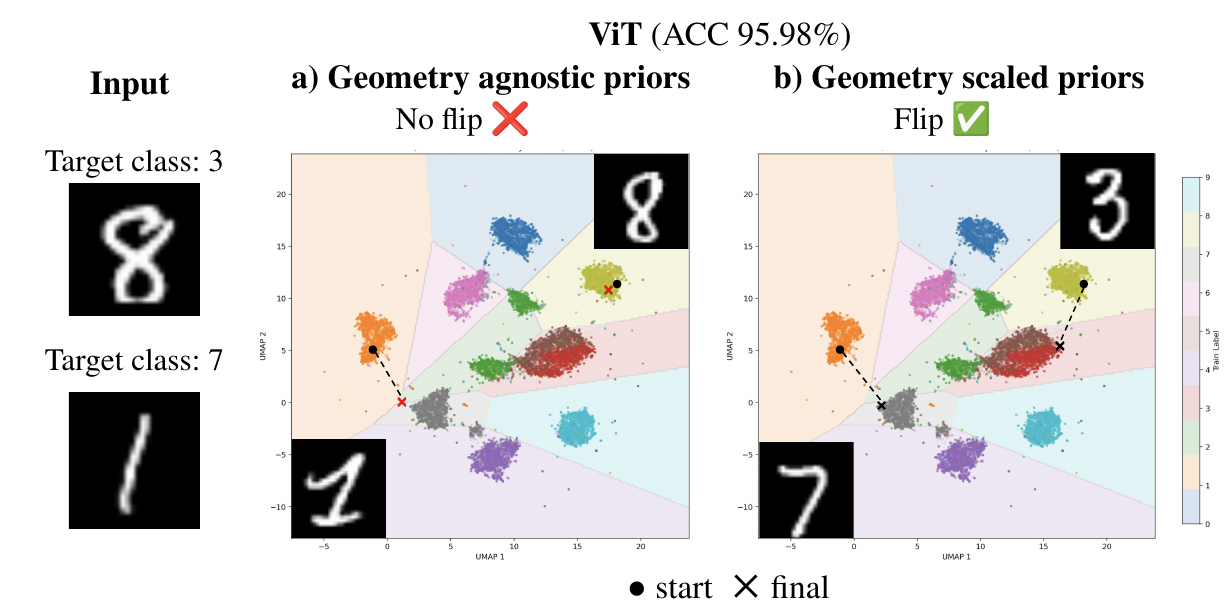}
    \caption{\textbf{Illustrative case study: geometry priors can improve local counterfactual search within the same model.} Two MNIST inputs are shown, each evaluated on the same pretrained model under the same optimization budget. For both examples, the baseline search does not reach a prediction change, whereas geometry-scaled priors successfully reach the specified target classes (3 and 7). Retrieved target class neighbors ground the final representation-space endpoints in input space. UMAP panels are shown for visualization only.}
\label{fig:usecase}
\end{figure}

\Cref{fig:usecase} compares the baseline search with a geometry-informed variant on two MNIST inputs under identical settings. 
For the 8$\rightarrow$3 example, the baseline fails to change the prediction and terminates farther from target-class examples (target-NN = 11.32), 
whereas the geometry-informed variant reaches a prediction change and ends closer to target support (target-NN = 6.06). 
To make the endpoints interpretable, the nearest target class training example is retrieved for each final point reached by the search. 
These retrieved images are \emph{not} counterfactuals themselves, but examples of the target class region approached by the trajectory. 
UMAP visualizations are for illustration only and do not preserve exact geometry; all quantitative results are computed in the original representation space. 
Additional examples appear in Appendix~\ref{app:usecases}. 

This case study highlights the value of geometry-informed counterfactual methods that account for the interaction between decision boundaries and local data support.

\section{Conclusion}

This work studies counterfactual behavior as a property of the interaction between decision-boundary proximity and local data support. 
Across diverse domains, models with similar predictive performance can differ substantially in whether nearby prediction changes are feasible, how far representations must move, and the required optimization effort. 
A consistent picture emerges: counterfactual behavior is governed by the interaction between decision-boundary proximity and local data support. 
Prediction changes are easiest when representations lie close to a boundary and near well-supported target regions, and become harder when either the boundary is distant or local support is weak. 
These quantities jointly explain variation in counterfactual success and distance across models and datasets.
Under fixed embeddings, modifying only the classifier head shifts boundary placement relative to the data, systematically altering behavior while leaving predictive performance largely unchanged. 
This sensitivity indicates that counterfactual behavior is not invariant across functionally similar models, but depends critically on classifier boundary placement relative to local data structure.
These findings establish counterfactual behavior as a distinct axis for evaluating learned models and suggest that counterfactual methods and recourse systems should account for the interaction between boundary proximity and local data support. 
Extending this perspective to end-to-end nonlinear models and developing geometry-aware counterfactual objectives remain important directions for future work. 
Limitations are discussed in Appendix~\ref{app:limitations}.

{
\small
\bibliography{bib}
\bibliographystyle{plain}
}
\newpage
\appendix

\section{Experimental Details}

\subsection{Dataset Details \& Statistics}

\paragraph{Shapes.}
Shapes~\cite{rudman-etal-2025-forgotten} is a synthetic image classification dataset containing simple geometric objects on a uniform background. 
Shape type, size, position, and color are systematically controlled, yielding low intra-class variation. 
It provides a useful controlled setting for studying counterfactual search and representation geometry.

\paragraph{MNIST.}
MNIST~\cite{lecun1998mnist} consists of grayscale $28 \times 28$ images of handwritten digits from 0 to 9. 
The dataset is visually simple but includes natural variation in handwriting style and stroke shape. 
It serves as a standard benchmark between synthetic and more complex real-image settings.

\paragraph{Chest X-ray (Pneumonia).}
The Chest X-ray dataset~\cite{Kermany2018} contains frontal chest radiographs labeled as pneumonia or normal. 
Visual differences between classes are subtle and can be affected by imaging variability.
This makes it a challenging medical-imaging setting for counterfactual analysis.

\paragraph{IMDb.}
IMDb~\cite{maasimdb} contains $50{,}000$ movie reviews evenly split between positive and negative sentiment. 
The task is defined over high-dimensional language representations capturing lexical and semantic structure. 
It provides a text-domain benchmark for studying whether counterfactual behavior varies across language encoders with similar predictive performance.

\paragraph{MM-IMDb.}
MM-IMDb~\cite{aravelo2017}contains approximately $15{,}600$ movie posters paired with plot summaries. 
The original multi-label task is restricted here to binary classification for controlled comparison. 
It enables evaluation across unimodal image models, unimodal text models, pretrained multimodal encoders, and simple fusion methods.

\subsection{Training}
For each dataset, a linear classifier is trained on top of frozen pretrained embeddings.
All encoders remain frozen throughout training to isolate the effects of representation geometry.

Classifier heads are trained with cross-entropy loss using Adam.
Unless otherwise specified, training uses a learning rate of $10^{-3}$,
batch size $128$, and at most $100$ epochs. 
Early stopping is based on validation accuracy.

For experiments with multiple classifier heads, 
the encoder and dataset are held fixed while heads are retrained from independent random initializations. 
For controlled boundary-shift experiments, the $\ell_2$ weight-decay parameter is varied while all other training settings remain fixed. 
Hyperparameters are kept consistent within each dataset to enable fair comparison across encoders and heads.

No stochastic image augmentation is applied to avoid introducing additional variability in the representation geometry across models.
For torchvision-based vision encoders, images are resized to $224\times224$, converted to RGB, transformed to tensors, and normalized using ImageNet channel statistics. 
For processor-based encoders (e.g., DINOv2, CLIP, SigLIP),
inputs are preprocessed using the model’s default pretrained image processor to ensure compatibility with the original training distribution.

\subsection{Counterfactual Optimization}

All main experiments use the same constrained local search procedure in representation space. The goal is not to compute globally optimal counterfactuals, but to provide a standardized probe of whether a nearby prediction change can be reached under fixed locality and optimization budgets. This matches the evaluation protocol used in the main paper. 

\paragraph{Objective.}
Let $z_0$ denote the original embedding with predicted class $y=f(z_0)$. The target class $y'$ is defined as the highest-scoring alternative class under the classifier:
\[
y'=\arg\max_{c \neq y} h(z_0)_c,
\]
where $h(z)$ denotes the classifier logits. Search minimizes the target class objective
\[
\ell(z_t) = -\, h(z_t)_{y'},
\]
which increases the target class logit and encourages a prediction change toward $y'$.

\paragraph{Projected Updates.}
Starting from $z_0$, iterative updates are given by
\[
z_{t+1}
=
\Pi_{\|z-z_0\|_2 \le \tau}
\left(
z_t-\eta \nabla_{z_t}\ell(z_t)
\right),
\]
where $\eta$ is the step size, $\tau$ is the trust-region radius, and $\Pi$ denotes Euclidean projection onto the $\ell_2$ ball centered at $z_0$. The projection enforces a fixed locality constraint so that all search trajectories remain within the same neighborhood of the original embedding.

\paragraph{Stopping Rule.}
Optimization terminates when either:
(i) the predicted label changes, i.e.,
\[
f(z_t)\neq f(z_0),
\]
or (ii) a maximum iteration budget $T$ is reached.

\paragraph{Shared Hyperparameters.}
Within each dataset, the parameters $\eta$, $\tau$, and $T$ are fixed and shared across all compared models. No geometry-aware priors, density terms, or model-specific tuning are used in the main protocol.
Geometry-aware priors are only introduced for the use-case in \cref{par:priors}.

\subsection{Compute Resources}\label{app:compute}

All experiments were conducted on an academic high-performance computing cluster.
The main compute cost was feature extraction from pretrained encoders. 
After embeddings were cached, linear probe training, SVM baselines, and downstream analyses were relatively lightweight, typically requiring minutes per run.

Total compute for the reported experiments was on the order of tens of GPU-hours, 
together with additional CPU-hours for probe training, search sweeps, and analysis,
excluding preliminary and discarded pilot runs.

\section{Additional Results}

\subsection{Loss-matched Model Comparisons}\label{app:loss}

To complement the accuracy-matched comparisons in the main text, this section repeats the paired-model analysis using validation cross-entropy (CE) loss.
For each dataset, we identify the pair of encoders with the smallest validation-loss gap and compare their counterfactual behavior.

\begin{table}[htb]
\centering
\small
\caption{\textbf{Closest-validation-loss encoder pairs within each dataset.}
Models with nearly identical validation cross-entropy can still exhibit large differences in counterfactual success (CF-Suc), counterfactual distance (CF-Dist), and optimization effort (OptEff). Reported values show mean differences with standard deviation across five seeds.}
\begin{tabular}{l l l r r r r}
\toprule
Dataset & Model A & Model B & CE Gap & CF-Suc Gap & CF-Dist Gap & OptEff Gap \\
\midrule
Shapes    
& ResNet50 & ViT     
& $0.08 \pm 0.009$ 
& $0.53 \pm 0.009$ 
& $0.38 \pm 0.007$ 
& $56.55 \pm 1.720$ \\

IMDB      
& BERT & RoBERTa 
& $0.01 \pm 0.004$ 
& $0.19 \pm 0.009$ 
& $0.49 \pm 0.007$ 
& $43.75 \pm 1.409$ \\

MNIST     
& ResNet50 & ViT     
& $0.06 \pm 0.008$ 
& $0.65 \pm 0.002$ 
& $0.23 \pm 0.001$ 
& $46.72 \pm 0.054$ \\

ChestXray 
& ResNet50 & ViT     
& $0.02 \pm 0.018$ 
& $0.27 \pm 0.044$ 
& $0.16 \pm 0.035$ 
& $29.39 \pm 3.096$ \\

MM-IMDb   
& CLIP     & SigLIP2 
& $0.02 \pm 0.002$ 
& $0.15 \pm 0.004$ 
& $0.11 \pm 0.003$ 
& $6.34 \pm 1.047$  \\
\bottomrule
\end{tabular}
\label{tab:main_gaps_nll}
\end{table}

Across datasets, small predictive-loss gaps can still coincide with substantial differences in counterfactual behavior. 
This is consistent with the main-text conclusion that counterfactual behavior is not fully characterized by standard predictive metrics alone.

\subsection{Comparison to Maximum-margin Linear Probes}\label{app:svm}

\begin{figure}[t]
    \centering
    \includegraphics[width=0.8\linewidth]{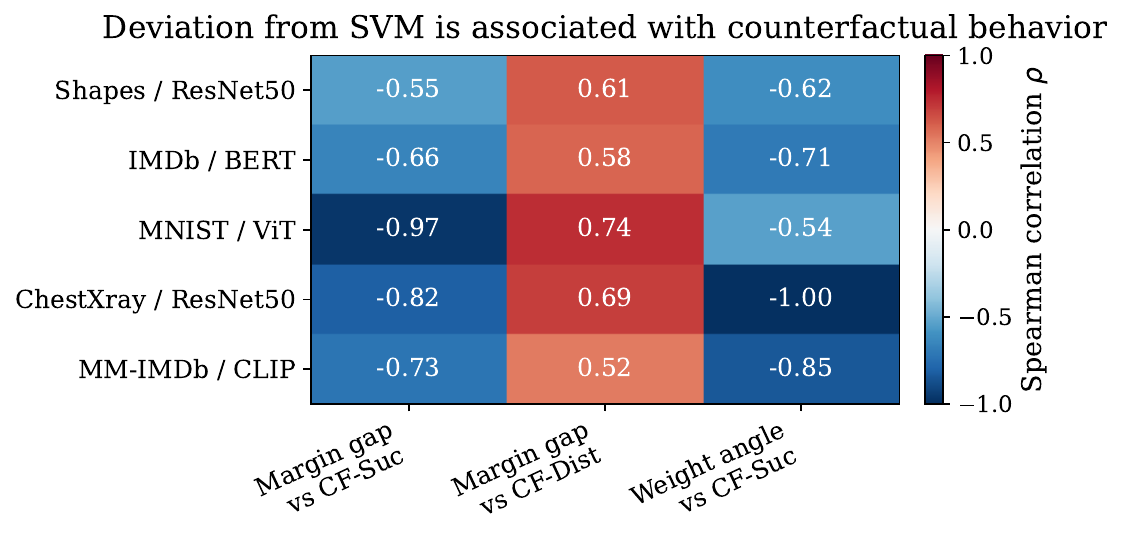}
    \caption{\textbf{Deviation from maximum-margin probes is associated with counterfactual behavior.}
    Spearman correlations between selected SVM-deviation metrics and counterfactual outcomes across datasets. Larger margin-gap or weight-angle deviation from the corresponding linear SVM is generally associated with lower CF-Suc and larger CF-Dist, supporting the role of classifier boundary placement in local counterfactual behavior.}
    \label{fig:svm_heatmap}
\end{figure}

To assess whether probe-to-probe variation is arbitrary or reflects systematic boundary placement, 
each learned probe is compared to a linear SVM trained on the same frozen embeddings. 
The SVM provides a canonical maximum-margin separator in the shared representation space.
\Cref{fig:svm_heatmap} reports correlations between selected deviation metrics and counterfactual outcomes across datasets. 
Margin gap measures the difference in average signed margin relative to the SVM, 
while weight angle measures the difference in orientation between the probe and SVM weight vectors.
Across datasets, probes closer to the SVM solution tend to exhibit better counterfactual behavior: 
higher CF-Suc and lower CF-Dist. On MNIST, for example, margin gap is strongly associated with both lower CF-Suc 
($\rho=-0.97$) and higher CF-Dist ($\rho=0.74$).

These results suggest that probe variability is structured:
even within fixed embeddings, boundary placement relative to a maximum-margin reference substantially influences local counterfactual behavior.
\subsection{Classifier heads Variation}

Table~\ref{tab:h2_appendix_ranges} shows that retraining classifier heads on the same frozen embeddings can substantially change counterfactual behavior while often leaving accuracy nearly unchanged. 
The effect varies across datasets and encoders: 
some settings exhibit little head-induced variation, whereas others show large CF-Suc shifts under minor accuracy changes. 
This suggests that representation geometry constrains the range of possible outcomes, 
but does not uniquely determine the behavior realized by the classifier.

\begin{table*}[htb]
\centering
\small
\caption{
\textbf{Variation in counterfactual behavior under fixed representations}. Reported ranges show the maximum change in predictive accuracy and counterfactual success across independently retrained classifier heads on the same frozen embeddings. Large counterfactual shifts can arise despite minor accuracy changes.}
\begin{tabular}{l l r r}
\toprule
Dataset & Encoder & $\Delta$ACC Range & $\Delta$CF-Suc Range \\
\midrule
Shapes    & DINOv2     & 0.00 & 0.12 \\
Shapes    & ResNet50   & 0.00 & 0.13 \\
Shapes    & ViT        & 0.00 & 0.11 \\
\midrule
IMDB      & BERT       & 0.00 & 0.33 \\
IMDB      & DistilBERT & 0.01 & 0.19 \\
IMDB      & RoBERTa    & 0.01 & 0.10 \\
\midrule
MNIST     & DINOv2     & 0.03 & 0.26 \\
MNIST     & ResNet50   & 0.01 & 0.12 \\
MNIST     & ViT        & 0.01 & 0.22 \\
\midrule
ChestXray & DINOv2     & 0.14 & 0.36 \\
ChestXray & ResNet50   & 0.02 & 0.22 \\
ChestXray & ViT        & 0.00 & 0.06 \\
\midrule
MM-IMDb   & DINOv2+BERT         & 0.03 & 0.40 \\
MM-IMDb   & DINOv2+DistilBERT   & 0.02 & 0.42 \\
MM-IMDb   & DINOv2+RoBERTa      & 0.01 & 0.34 \\
MM-IMDb   & ResNet50+BERT       & 0.00 & 0.17 \\
MM-IMDb   & ResNet50+DistilBERT & 0.01 & 0.12 \\
MM-IMDb   & ResNet50+RoBERTa    & 0.02 & 0.30 \\
MM-IMDb   & ViT+BERT            & 0.02 & 0.48 \\
MM-IMDb   & ViT+DistilBERT      & 0.02 & 0.28 \\
MM-IMDb   & ViT+RoBERTa         & 0.02 & 0.59 \\
MM-IMDb   & CLIP                & 0.01 & 0.17 \\
MM-IMDb   & SigLIP2             & 0.01 & 0.13 \\
\bottomrule
\end{tabular}
\label{tab:h2_appendix_ranges}
\end{table*}

\clearpage
\subsection{Case studies} \label{app:usecases}

We show qualitative examples across vision, medical, text, and multimodal settings.
Each compares geometry-agnostic search with geometry-aware priors under identical budgets.
Geometry-agnostic search often fails,
while geometry-aware priors guide trajectories to supported target regions and achieve successful prediction changes.
Text is trimmed for display, and retrieved examples illustrate nearby target regions rather than the counterfactuals themselves.

\begin{figure}[h]
    \centering
    \includegraphics[width=0.8\linewidth]{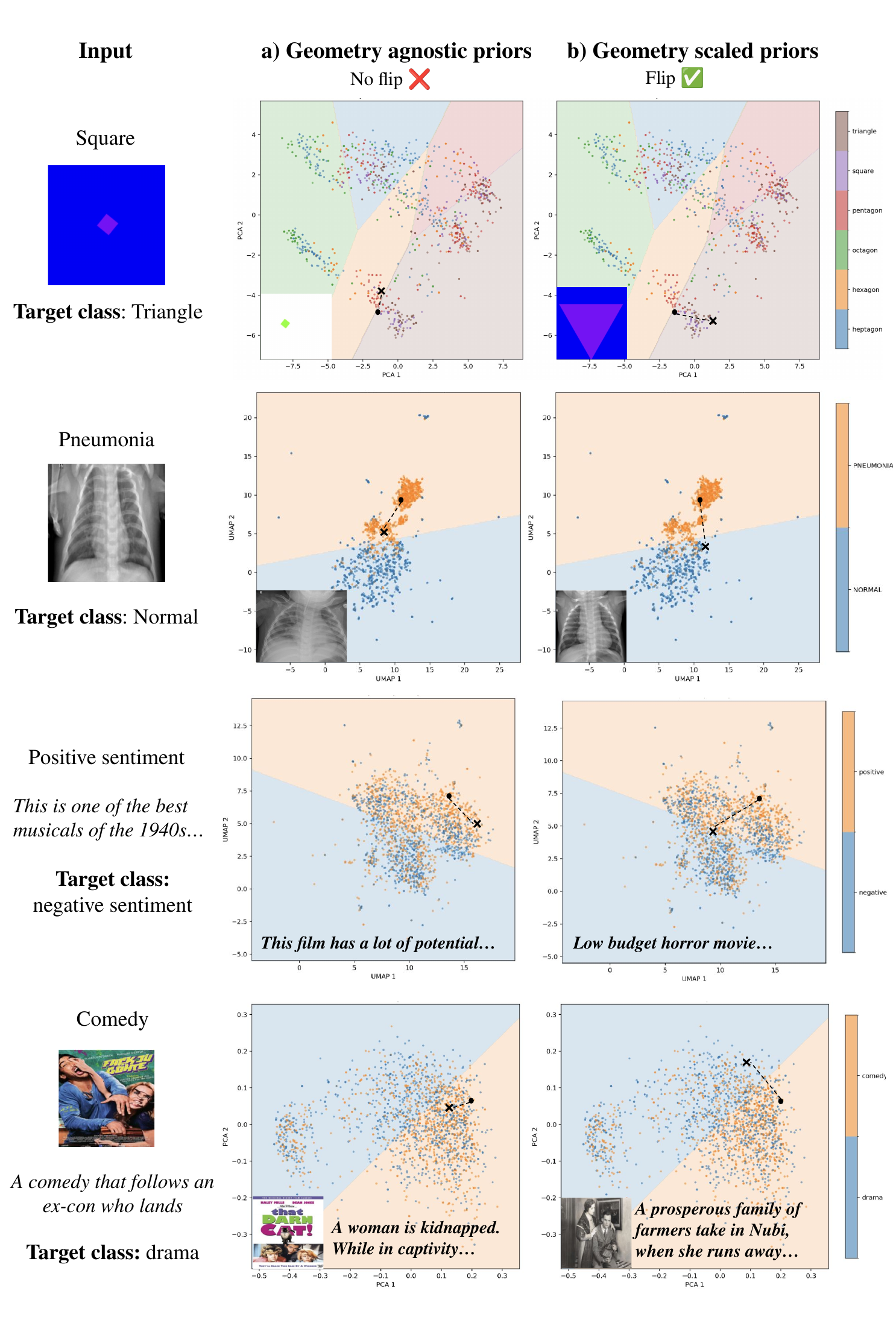}
\caption{\textbf{Geometry-aware priors improve counterfactual search across modalities.}
Across vision, medical, text, and multimodal examples, geometry-agnostic search often fails (no flip), while geometry-scaled priors guide trajectories toward target class regions and achieve successful prediction changes. Low-dimensional projections exhibit weak class separation. Retrieved images/text illustrate nearby target regions, not counterfactuals.}
\label{fig:placeholder}
\end{figure}
\clearpage

\section{Robustness Analyses}

\subsection{Density Estimation}
\label{app:density}

Local data support is approximated using the $k$-nearest neighbor (k-NN) radius in representation space. 
Since this is only a proxy for the underlying data distribution, 
we evaluate whether the main conclusions depend on the neighborhood size.

\begin{figure}[h]
    \centering
    \includegraphics[width=0.9\linewidth]{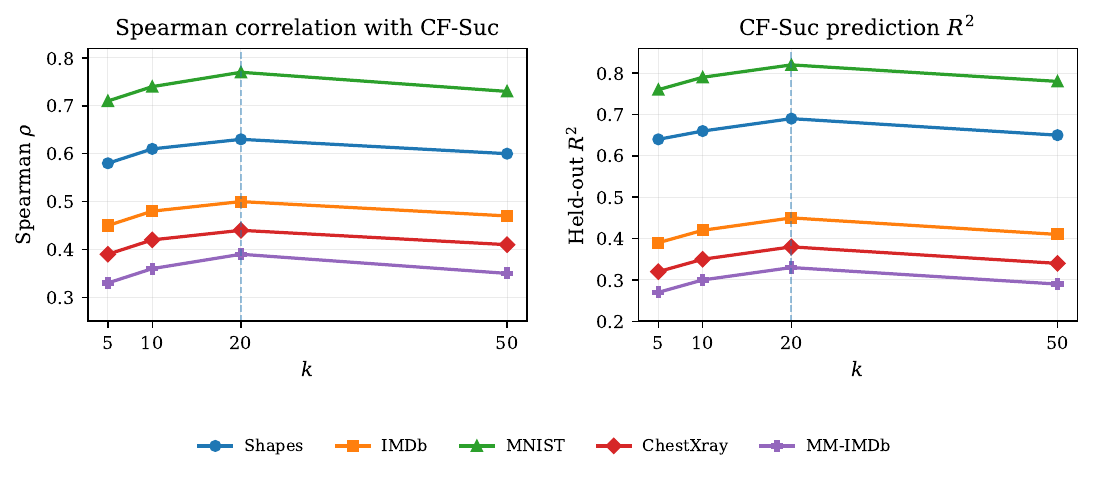}
    \caption{\textbf{Sensitivity to neighborhood size.}
    Spearman correlation between local data support and CF-Suc (left), and held-out $R^2$ for CF-Suc prediction using decision-boundary proximity and local data support (right), across $k \in \{5,10,20,50\}$. Results vary mildly across the tested neighborhood sizes, with $k=20$ strongest in this sweep.}
    \label{fig:knn_k_robustness}
\end{figure}

\paragraph{Sensitivity to neighborhood size.}
We vary $k \in \{5,10,20,50\}$ and measure (i) the Spearman correlation between $r_k(z)$ and CF-Suc, 
and (ii) the held-out $R^2$ of the post hoc models in Section~5 that use decision-boundary proximity and local data support as features. 
Results are summarized in \cref{fig:knn_k_robustness}.

Across datasets, both quantities vary only mildly across neighborhood sizes. 
The curves remain close over the tested range, with $k=20$ giving the strongest values in this sweep. 
These results suggest that the main trends are not sensitive to the precise choice of $k$ within the tested range.

\paragraph{Scale normalization.}
To account for differences in representation scale, we evaluate normalized variants of $r_k(z)$, including (i) standardization by global feature variance and (ii) normalization by the median pairwise distance within each dataset. Both transformations leave correlations and $R^2$ values effectively unchanged (differences $<0.02$), suggesting that the observed relationships are not driven by scale.

\subsection{Optimization Settings}

One concern is that the observed differences are artifacts of the search procedure rather than properties of the models. To test this, we vary the optimizer (SGD, Adam, AdamW), step size $\eta$, trust-region radius $\tau$, and iteration budget one at a time from a fixed default configuration.

Across datasets, relative model rankings remain broadly stable under these changes, although sensitivity varies by parameter. Changing the optimizer has the smallest effect, while modifying step size or trust-region radius produces larger shifts in rank correlation. Reducing the iteration budget also lowers agreement relative to more permissive budgets, whereas larger budgets yield similar rankings. These patterns indicate that the reported comparisons are not tied to a single search configuration, while also showing that overly restrictive settings can weaken stability.

\begin{figure}[h]
    \centering
    \includegraphics[width=0.9\linewidth]{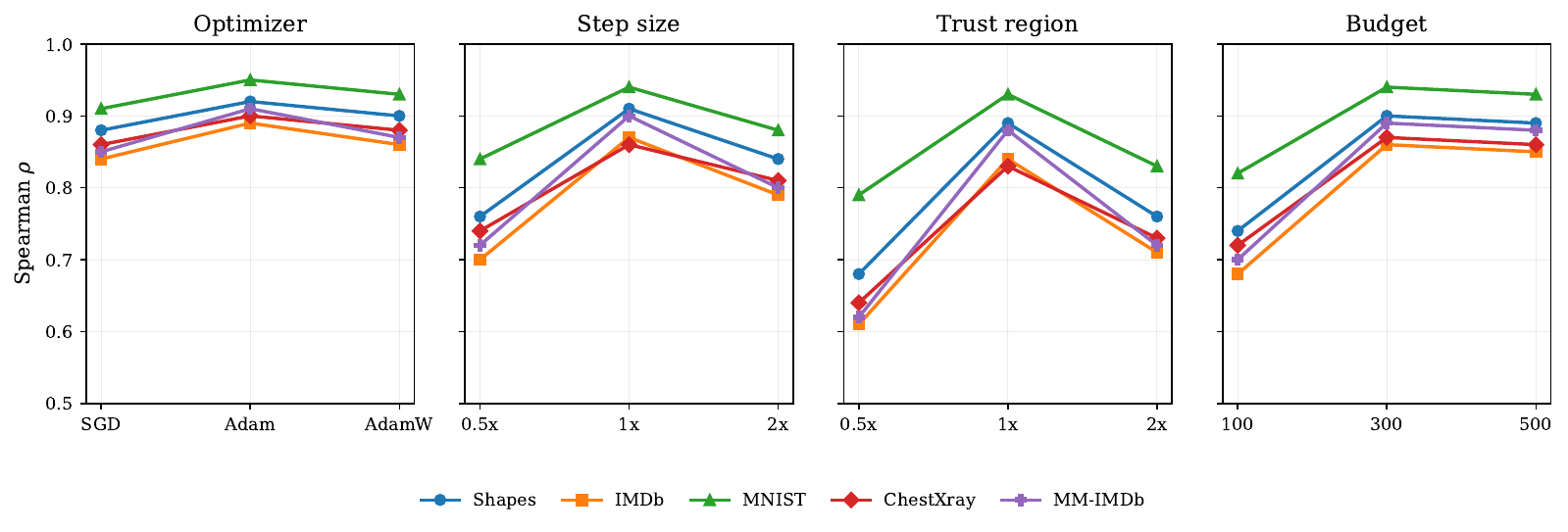}
    \caption{\textbf{Model rankings are broadly stable across optimization settings.}
    Spearman rank correlation of model-level CF-Suc between the default configuration and alternative settings. Optimizer changes have the smallest effect, step size and trust-region changes have larger effects, and larger iteration budgets produce similar rankings. Overall, the relative ordering of models is largely preserved across settings.}
    \label{fig:rank}
\end{figure}

\section{Limitations \& Future Work} \label{app:limitations}

This study has several limitations. 
First, the experiments focus on frozen pretrained encoders paired with linear classifier heads. 
This design is useful for separating representation geometry from classifier boundary geometry, 
but some findings may not fully extend to end-to-end nonlinear models in which representations and decision boundaries are learned jointly.
Second, counterfactual behavior is evaluated using a specific local search protocol with fixed trust-region constraints, optimization budgets, and target class selection rules. Although model rankings are robust across several optimizer and hyperparameter choices, the reported metrics remain conditional on this search formulation and may differ under alternative counterfactual procedures.
Third, the analysis is conducted primarily in representation space rather than input space.
A successful embedding-space prediction change does not by itself guarantee that a corresponding input is semantically plausible, human interpretable,
or feasible under domain-specific feature constraints.
The nearest-neighbor grounding examples are illustrative only and are not substitutes for valid input-space counterfactuals.
Finally, local data support is estimated using finite-sample proxies such as class-conditional $k$-nearest-neighbor distances and related alternatives. These quantities capture useful local structure, but they are imperfect proxies for the underlying data distribution and may be sensitive to sample size, class imbalance, and embedding anisotropy.

Thus, promising directions for future work include extending this perspective to end-to-end nonlinear models, developing richer input-space counterfactual objectives, and designing search procedures that adapt to measured representation geometry.

\section{Impact Statement}

This paper studies counterfactual behavior in learned models, with potential implications for interpretability, auditing, recourse, and model selection. Understanding when prediction changes are feasible and locally supported may help improve transparency and reliability in decision-making systems. As with related work on model behavior and robustness, similar analyses could also be used to probe model weaknesses or decision boundaries.

\end{document}